\documentclass{article}
\usepackage{amsmath,graphicx}

\usepackage[english]{babel}

\usepackage{ifpdf}

\usepackage{cite} % Orders citations.
\usepackage{url}
\usepackage{hyperref}
\usepackage{color}
\usepackage{pgf, tikz, pgfplots}
\usetikzlibrary{shapes, arrows, automata, plotmarks}
\usetikzlibrary{calc,hobby,decorations}
\usepackage{tikz-cd}

\usepackage{stfloats}
\usepackage{amsfonts, amssymb, amsthm}
\usepackage{algorithm,algpseudocode}
	\algnewcommand{\LeftComment}[1]{\Statex \(\triangleright\) #1}

\usepackage{subfloat}
\usepackage{enumerate}
\usepackage{multirow}
\usepackage{rotating}
\usepackage{subcaption}
\usepackage{needspace}

\input{./latex_resources/mySymbol.sty}
\input{./latex_resources/pennColors.sty}

\usepackage{./latex_resources/spconf}

\newtheorem{example}{\hspace{0pt}\bf Example}

\newtheorem{theorem}{\hspace{0pt}\bf Theorem}

\newtheorem{definition}{\hspace{0pt}\bf Definition}

\ninept

\title{ Non Commutative Convolutional Signal Models in Neural Networks: Stability to Small Deformations }

\name{Alejandro Parada-Mayorga$^{\ast}$, Landon Butler$^{\dagger}$, and Alejandro Ribeiro$^{\ast}$
\thanks{This document is a conference version of~\cite{algnn_nc_j} which was recently published in IEEE-TSP. The second author acknowledges support by the NSF Graduate Research Fellowship under Grant No. DGE-2146752.}
}
\address{University of Pennsylvania$^{\ast}$\\ University of California, Berkeley$^{\dagger}$}

\begin{document}
\maketitle
\begin{abstract}

In this paper we discuss the results recently published in~\cite{algnn_nc_j} about algebraic signal models (ASMs) based on non commutative algebras and their use in convolutional neural networks. Relying on the general tools from algebraic signal processing (ASP), we study the filtering and stability properties of non commutative convolutional filters. We show how non commutative filters can be stable to small perturbations on the space of operators. We also show that although the spectral components of the Fourier representation in a non commutative signal model are associated to spaces of dimension larger than one, there is a trade-off between stability and selectivity similar to that observed for commutative models. Our results have direct implications for group neural networks, multigraph neural networks and quaternion neural networks, among other non commutative architectures. We conclude by corroborating these results through numerical experiments.  

\end{abstract}
\begin{keywords}
Non commutative convolutional architectures, Algebraic Neural Networks (AlgNNs), algebraic signal processing (ASP), non commutative algebras, non commutative operators.
\end{keywords}
%
%
%

%%%%%%%%%%%%%%%%%%%%%%%%%%%%%%%%%%%%%%%%%%%%%%%
%%%%%%%%%%%%%%%%% SECTION: INTRODUCTION %%%%%%%%%%%%%%%
%%%%%%%%%%%%%%%%%%%%%%%%%%%%%%%%%%%%%%%%%%%%%%%

%!TEX root = ../conf_paper.tex

%%%%%%%%%%%%%%%%%%%%%%%%%%%%%%%%%%%%%%%%%%%%%%%
%%%%%%%%%%%%%%%%% INTRODUCTION %%%%%%%%%%%%%%%%%%%%%
%%%%%%%%%%%%%%%%%%%%%%%%%%%%%%%%%%%%%%%%%%%%%%%

\section{Introduction}
\label{sec:intro}

Convolutional signal models have become ubiquitous tools in machine learning.  Understanding their fundamental properties plays a central role in explaining convolutional neural networks' good performance and limitations in different domains. One such property is that of stability to deformations. Previous works have shown that the good performance of convolutional architectures can be explained in part by the fact that for the same discriminability, a convolutional network is more stable to deformations than the corresponding convolutional filters~\cite{parada_algnn, parada_algnnconf, mallat_ginvscatt,fern2019stability,agg_gnn_j}. These results apply in commutative scenarios such as graph filters and graph neural networks~\cite{fern2019stability}, Euclidean filters and traditional convolutional neural networks~\cite{mallat_ginvscatt}, graphon filters and graphon neural networks~\cite{gphon_pooling_c,gphon_pooling_j}, and commutative group filters and their associated group neural networks~\cite{parada_algnn, parada_algnnconf}. However, the question of stability has remained open for non commutative signal models and architectures such as multigraph neural networks~\cite{msp_j,msp_icassp2023}, Lie group neural networks~\cite{lga_j,lga_icassp}, quaternion neural networks~\cite{parcollet2019quaternion}, and quiver neural networks~\cite{parada_quiversp}.

In this work we develop a formal description of non commutative signal models as a particular instantiation of an algebraic signal model with a non commutative algebra. We leverage this algebraic representation to study the stability of non commutative convolutional filters and their associated neural networks. We show that although the spectral representations of non commutative convolutional filters are described by spaces whose dimension is larger than one, there is still a trade-off between discriminability and stability, similar to the one observed in commutative scenarios. Additionally, we show that for the same level of discriminability, the neural networks are more stable than the corresponding non commutative convolutional filters, under the notion of deformation and stability used in~\cite{parada_algnn, parada_algnnconf}. We also prove that the stability bounds for non commutative convolutional networks are a scaled version of the bounds derived for the filters. This implies that the networks inherit the stability properties from the filters as a consequence of the attributes of the pointwise nonlinearities and the pooling operators. We conclude with numerical experiments to validate our results.

%%%%%%%%%%%%%%%%%%%%%%%%%%%%%%%%%%%%%%%%%%%%%%%
%%%%%%%%%%%%% NON COMMUTATIVE SIGNAL MODELS %%%%%%%%%%%%%%%
%%%%%%%%%%%%%%%%%%%%%%%%%%%%%%%%%%%%%%%%%%%%%%%

%!TEX root = ../conf_paper.tex

%%%%%%%%%%%%%%%%%%%%%%%%%%%%%%%%%%%%%%%%%%%%%%%%%%%
%%% %%%%%%%%%%%%  S   E   C   T   I   O   N   %%%%%%%%%%%%%%%%%%%%%%%%
%%%%%%%%%%%%%%%%%%%%%%%%%%%%%%%%%%%%%%%%%%%%%%%%%%%

\section{Non Commutative Convolutional Signal Models} \label{sec_alg_filters}

% %%----------------------------------------
% %%--------------- FIGURE ------------------
% %%----------------------------------------

% \begin{figure}
%     \centering
%     \input{./figures/50_asp}
%     \caption{Non commutative algebraic signal model. The algebraic filters $th$ and $ht$ are \textit{realized} physically in $\text{End}(\ccalM)$ to process the signals $\bbx$ which are modeled as elements of $\ccalM$. Adapted from~\cite{algnn_nc_j}.}
%     \label{fig:my_ncASP}
% \end{figure}

% %%------------ End of FIGURE -------------

The general notion of a convolutional signal model in the algebraic sense was first introduced by Puschel in~\cite{algSP0, algSP1, algSP2, algSP3} and is known as \textit{algebraic signal processing (ASP)}. In this context, any convolutional signal model is given by a triplet $(\ccalA, \ccalM, \rho)$, where $\ccalA$ is a unital associative algebra, $\ccalM$ is a vector space and $\rho$ is a homomorphism from $\ccalA$ to the space of endomorphisms of $\ccalM$, $\text{End}(\ccalM)$. The elements in $\ccalA$ are the filters, signals are the elements of $\ccalM$ and $\rho: \ccalA \to \text{End}(\ccalM)$ instantiates the abstract filters in $\ccalA$ into concrete operators acting on $\ccalM$.

Most applications of the ASP paradigm are associated with scenarios where the filters are determined by commutative operators~\cite{dsp_graphsmoura,puschel_setfunc,puschel_lattice,gphon_pooling_c,gphon_pooling_j}. However, it is possible to rely on the same framework to introduce convolutional non commutative models choosing $\ccalA$ as a non commutative algebra~\cite{algnn_nc_j}. Then, any non commutative convolutional model to be considered in this paper can be described by the triplet $(\ccalA,\ccalM,\rho)$ where $\ccalA$ is a non commutative algebra. If $\bbx\in\ccalM$ is a signal, we refer to the signal $\bby = \rho(a)\bbx$ as the filtered version of $\bbx$ by means of the filter $a\in\ccalA$. More specifically, we say that $\bby$ is the result of performing a convolution between the filter $a\in\ccalA$ and the signal $\bbx\in\ccalM$.

To simplify the representation of the elements in $\ccalA$, we leverage the concept of a \textit{generator set} of an algebra. We say that the set $\ccalG\subset\ccalA$ generates the algebra $\ccalA$ if any element of $\ccalA$ can be written as a polynomial of the elements in $\ccalG$. For instance, if $\ccalG = \{ g_1, g_2 \}$, then any element $a\in\ccalA$ can be written as $a = p(g_1,g_2)$, where $p$ is a polynomial with two independent variables. If $\ccalA$ is non commutative, the elements of $\ccalG$ do not commute. The image of a generator, $g\in\ccalG$, under $\rho$ is known as a \textit{shift operator}, which we denote by $\bbS = \rho(g)$. Since $\rho$ is a homomorphism, it is a linear map that preserves the products in $\ccalA$. Therefore, the image of any filter in $\ccalA$ can be written as a polynomial of the shift operators. Additionally, the coefficients of such polynomials are the same used for the filter in $\ccalA$. For instance, if $\ccalG = \{ g_1, g_2 \}$ and $a = p(g_1,g_2)$, we have $\rho(a)=p(\bbS_1, \bbS_2)$ where $\bbS_i = \rho(g_i)$.

%%%%%%%%%%%%%%%%%%%%%%%%%%%%%%%%%%%%%%%%%%%%%%%
%%%%%%%%%%%%% STABILITY OF ALGEBRAIC FILTERS%%%%%%%%%%%%%%%
%%%%%%%%%%%%%%%%%%%%%%%%%%%%%%%%%%%%%%%%%%%%%%%

%!TEX root = ../conf_paper.tex

%%%%%%%%%%%%%%%%%%%%%%%%%%%%%%%%%%%%%%%%%%%%%%%%%%%
%%% %%%%%%%%%%%%  S   E   C   T   I   O   N   %%%%%%%%%%%%%%%%%%%%%%%%
%%%%%%%%%%%%%%%%%%%%%%%%%%%%%%%%%%%%%%%%%%%%%%%%%%%

\section{Fourier Decompositions of Convolutional Non Commutative Filters} \label{sec_spectral_rep}

% %%------------------------------------------
% %%---------------- FIGURE ------------------
% %%------------------------------------------

% \begin{figure}
% 	\centering \input{./figures/fig_1_tikz_source.tex}
% \caption{ 
% General notion of Fourier decomposition in an ASM. The homomorphism, $\phi_i$, determine the Fourier representation of the filter, $a\in\ccalA$, on the $i$-th frequency. Since for any non commutative model $\text{dim}(\ccalU_i)>1$ for at least one $i$, $\phi_i (a)$ is a matrix polynomial function with independent variables $\phi_1 (g)$, where $g$ is a generator of $\ccalA$. Each space, $\ccalU_i$, is invariant under the action of $\rho(a)$ for any $a\in\ccalA$. Additionally, $\rho(a)\ccalU_i =\phi_{i}(a)\ccalU_i$. Adapted from~\cite{algnn_nc_j}. 
% }
% \label{fig_3}
% \end{figure}

% %%-------------- End of FIGURE ---------------

The notion of Fourier decomposition in ASP is rooted in the concept of \textit{decomposition in terms of irreducible subrepresentations}~\cite{algSP0,algnn_nc_j,parada_algnn}. To see this, let us recall a few basic facts about ASM using concepts from representation theory of algebras. First, we emphasize that for any ASM, $(\ccalA, \ccalM, \rho)$, the pair $(\ccalM, \rho)$ constitutes a representation of $\ccalA$ in the sense of~\cite{repthybigbook,repthysmbook}. Then, a representation $(\mathcal{U},\rho)$ of $\mathcal{A}$ is a \textit{subrepresentation} of  $(\mathcal{M},\rho)$ if $\mathcal{U}\subseteq\mathcal{M}$ and $\rho(a)u\in\mathcal{U}$ for all $u\in\mathcal{U}$ and $a\in\mathcal{A}$~\cite{algnn_nc_j}. A subrepresentation of $(\ccalM,\rho)$ can be obtained when one finds a subspace of $\ccalM$ invariant under the action of any operator $\rho(a)$. Using this concept, we introduce the essential concept of irreducible representation. We say that $(\mathcal{M},\rho)$, when $\ccalM\neq 0$, is \textit{irreducible} or simple if the only subrepresentations of $(\mathcal{M},\rho)$ are $(0,\rho)$ and $(\mathcal{M},\rho)$. Finally, we state the notion of decomposition of a representation as a sum of other representations. Let $(\ccalU_1, \rho_1)$ and $(\ccalU_1, \rho_2)$ be two representations of $\ccalA$. Then, we can obtain a new representation called the \textit{direct sum representation}, given by $(\ccalU_1 \oplus \ccalU_1, \rho)$, where $\rho(a)(u_1 \oplus u_2) \cong \rho_1 (a)u_1 \oplus \rho_2 (a)u_2 $.

With these concepts at hand, we present the definition of Fourier decomposition that we will use for the frequency characterization of convolutional non commutative filters.

%%---------------------------------------------------------------------------------
%%----------------------------- DEFINITION ---------------------------------------
%%---------------------------------------------------------------------------------

\begin{definition}[Fourier Decomposition~\cite{algnn_nc_j}]\label{def:foudecomp}
For an algebraic signal model $(\mathcal{A},\mathcal{M},\rho)$, we say that there is a spectral or Fourier decomposition if
\begin{equation}
(\mathcal{M},\rho)\cong\bigoplus_{(\mathcal{U}_{i},\phi_{i})\in\text{Irr}\{\mathcal{A}\}}(\mathcal{U}_{i},\phi_{i})
,
 \label{eq:foudecomp1}
\end{equation}
where the $(\mathcal{U}_{i},\phi_{i})$ are irreducible subrepresentations of $(\mathcal{M},\rho)$. Any signal $\mathbf{x}\in\mathcal{M}$ can be therefore represented by the map $\Delta$ given by
$
\Delta: \mathcal{M} \to \bigoplus_{(\mathcal{U}_{i},\phi_{i})\in\text{Irr}\{\mathcal{A}\}}\mathcal{U}_{i}
$, with
$
\mathbf{x}\mapsto \hat{\mathbf{x}}
,
$
known as the Fourier decomposition of $\mathbf{x}$. The projection of $\hat{\mathbf{x}}$ on $\mathcal{U}_{i}$ is the $i$-th Fourier component of $\bbx$ and is represented by $\hat{\mathbf{x}}(i)$.
\end{definition}

%%---------------------- End of DEFINITION ----------------------------------------

Although abstract, Definition~\ref{def:foudecomp} offers a consistent, general and rigorous description of the Fourier transform under any ASM, which is independent from the domain of the signals. Additionally, from Definition~\ref{def:foudecomp}, there are three fundamental observations in terms of the practicality of the Fourier decomposition and how to find it. First, the Fourier decomposition emphasizes a decomposition of the signals in terms of spaces that are invariant under the action of any $\rho(a)$. Second, such minimum invariant subspaces are given by the irreducible subrepresentations of $(\ccalM, \rho)$. Third, the restriction of $\rho(a)$ to each invariant subspace $\ccalU_i$ is determined by a homomorphism $\phi_i (a)$ acting from $\ccalA$ to $\text{End}(\ccalU_i)$. Additionally, the $\phi_i (a)$ determine the Fourier representation of $a$ in the $i$-th frequency component.

Note that Definition~\ref{def:foudecomp} is used for commutative and non commutative models alike. However, its instantiation in both scenarios have substantial differences. While for a commutative model $\text{dim}(\ccalU_i)=1$, in non commutative models it must be that $\text{dim}(\ccalU_i)>1$ for at least one $i$-th. This implies that the spectral representation of non commutative filters are characterized by matrix polynomials. To see this, let us consider the scenario where $\ccalA$ is non commutative with generator set $\ccalG =\{ g_1, g_2 \}$. Since for at least one of the $\ccalU_i$ we have $\text{dim}(\ccalU_i)>1$, the space of endomorphisms $\text{End}(\ccalU_i)$ is a space of matrices of size $\text{dim}(\ccalU_i)\times \text{dim}(\ccalU_i)$. Then, the Fourier representation of a filter $a = p(g_1, g_2)$ is given by $\phi_i (a) = p(\phi_i (g_1), \phi_i (g_2))$ which is a matrix polynomial where the independent variables are the matrices $\phi_i (g_1)$ and $\phi_i (g_2)$. The projection of $\bbx$ on $\ccalU_i$, $\hat{\bbx}(i)$, is the $i$-th Fourier component of $\bbx$.

As pointed out in~\cite{msp_j, msp_icassp2023}, when the dimension of $\ccalM$ is finite, the problem of finding the Fourier decomposition of an ASM translates into solving a joint block diagonalization problem involving the shift operators. To see this, consider that $\ccalA$ has an arbitrary number of generators, $m$, and let $(\ccalA, \ccalM, \rho)$ be an ASM with shift operators $\{ \bbS_{i}\}_{i=1}^{m}$, which have a joint block diagonalization given by
$
\bbS_i 
      =
       \bbU 
           \textnormal{diag}\left( 
                        \boldsymbol{\Sigma}_{1}^{(i)} , \ldots, \boldsymbol{\Sigma}_{\ell}^{(i)}
                       \right) 
       \bbU^{T}
       ,
$
with $\boldsymbol{\Sigma}_{j}^{(i)}\in\mbR^{p_j \times p_j}$, and $\bbU$ is orthogonal. Then, if we choose $d = \max_j \{ p_j \}$ and $\boldsymbol{\Lambda}_{i}\in \mbR^{d\times d}$, we have that the Fourier representation of a filter $p(\bbS_1, \ldots, \bbS_m)= \rho\left(p(g_1, \ldots, g_m)\right)$ is given by the matrix polynomial
\begin{equation}
	p
	  \left(
	        \boldsymbol{\Lambda}_1, \ldots, \boldsymbol{\Lambda}_m
	  \right)
	  :
	  \left( \mbR^{d\times d} \right)^m \rightarrow \mbR^{d\times d}
	  ,
\end{equation}
where $\left( \mbR^{d\times d} \right)^m$ is the $m$-times cartesian product of $\mbR^{d\times d}$. To clarify the concepts discussed above, we present the following example.

%%-------------------------------------
%%------------- EXAMPLE ---------------
%%-------------------------------------

\begin{example}\normalfont
Let $(\ccalA, \mbR^3, \rho)$ be a non commutative ASM where $\ccalA$ has generator set $\ccalG =\{ g_1 , g_ 2\}$ and where $(\ccalM, \rho)= (\ccalU_1, \phi_1)\oplus (\ccalU_2, \rho_2)$ with $\text{dim}(\ccalU_1)=1$ and $\text{dim}(\ccalU_2)=2$. Then, the Fourier representation of the filter $p(g_1,g_2)= g_1 + 5g_{1}g_{2}+g_{2}^{2}$ is given by the matrix polynomial $p(\boldsymbol{\Lambda}_1,\boldsymbol{\Lambda}_2)= \boldsymbol{\Lambda}_1 + 5\boldsymbol{\Lambda}_{1}\boldsymbol{\Lambda}_{2}+\boldsymbol{\Lambda}_{2}^{2}$ where $\boldsymbol{\Lambda}_i \in M_{2\times 2}$.
\end{example}

%%--------- End of Example ------------

%%%%%%%%%%%%%%%%%%%%%%%%%%%%%%%%%%%%%%%%%%%%%%%
%%%%%%%%%%%%% ALGEBRAIC NEURAL NETWORKS %%%%%%%%%%%%%%%
%%%%%%%%%%%%%%%%%%%%%%%%%%%%%%%%%%%%%%%%%%%%%%%

%!TEX root = ../conf_paper.tex

%
%%%%%%%%%%%%%%%%%%%%%%%%%%%%%%%%%%%%%%%%%%%%%%%%%%%%%%%%%%%%%%%%%%%%%%%%%%%%%%%%
%%%   S   E   C   T   I   O   N   %%%%%%%%%%%%%%%%%%%%%%%%%%%%%%%%%%%%%%%%%%%%%%
%%%%%%%%%%%%%%%%%%%%%%%%%%%%%%%%%%%%%%%%%%%%%%%%%%%%%%%%%%%%%%%%%%%%%%%%%%%%%%%%
%
\section{Non Commutative Convolutional Neural Networks}\label{sec_Algebraic_NNs}

%%------------------------------
%%---------  FIGURE  -----------
%%------------------------------

\begin{figure}
\centering% !TEX root = ../00_algnn_penn.tex

%----- Cividis color palette -----
\definecolor{my_cp_col1}{RGB}{0, 32, 81}
\definecolor{my_cp_col2}{RGB}{43, 68, 110}
\definecolor{my_cp_col3}{RGB}{105, 105, 112}
\definecolor{my_cp_col4}{RGB}{148, 143, 120}
\definecolor{my_cp_col5}{RGB}{202, 186, 106}
\definecolor{my_cp_col6}{RGB}{253, 234, 69}

\def \myfactor {0.6}
\def \unit  {\myfactor cm}

% Definition of a block for block diagrams
\tikzstyle{block} = [ rectangle,
                      minimum width = \unit,
                      minimum height = \unit,
                      fill = black,
                      draw = black,
                      text = black]

% Control the color and size of filter blocks                  
\tikzstyle{filter} = [block,
                      minimum width  = 3.0*\unit,
                      minimum height = 1.1*\unit,
                      fill=my_cp_col1]

% Control the color and size of nonlinearity blocks 
\tikzstyle{nonlinearity} = [ filter,
                             minimum width  = 3.0*\unit,
                             fill =my_cp_col3]
                             %fill = mygreen!15]

% Control the color and size of nonlinearity blocks 
\tikzstyle{pooling} = [ filter,
                             minimum width  = 3.0*\unit,
                             fill = my_cp_col6]
                             %fill = mygreen!15]

% Control separation between blocks
\def \deltainput     {( 0.0,-1.7)}
\def \deltaoutput    {( 0.0,-1.2)}
\def \deltalayer     {3.1}
\def \deltaconnector {1.45}
\def \deltasigma     {( 4, 0.0)}

% The text that is displayed inside of filter blocks
\def\one{\textcolor{my_cp_col6}{$\displaystyle{\mathbf{y}_{1}  = \rho_{1}(a_1)\,\mathbf{x}}$}}

\def\two{\textcolor{my_cp_col6}{$\displaystyle{\mathbf{y}_2  =  \rho_{2}(a_2)\,\mathbf{x}_{1}}$}}

\def\three{\textcolor{my_cp_col6}{$\displaystyle{\mathbf{y}_3  = \rho_{3}(a_3)\,\mathbf{x}_{2}}$}}

% The text that is displayed inside of nonlinearity blocks
\def\sigmaone{\textcolor{my_cp_col6}{$\displaystyle{\mathbf{z}_{1} = {\eta_{1}} \Big[\, \mathbf{y}_1 \, \Big]}$}}

\def\sigmatwo{\textcolor{my_cp_col6}{$\displaystyle{\mathbf{z}_{2} = {\eta_{2}} \Big[\, \mathbf{y}_2 \, \Big]}$}}

\def\sigmathree{\textcolor{my_cp_col6}{$\displaystyle{\mathbf{z}_{3} = {\eta_{3}} \Big[\, \mathbf{y}_3 \, \Big]}$}}

\def \proyone   {$\displaystyle{\mathbf{x}_{1} = {P_{1}} \Big[\, \mathbf{z}_1 \, \Big]}$}
\def \proytwo   {$\displaystyle{\mathbf{x}_{2} = {P_{2}} \Big[\, \mathbf{z}_2 \, \Big]}$}
\def \proythree   {$\displaystyle{\mathbf{x}_{3} = {P_{3}} \Big[\, \mathbf{z}_3 \, \Big]}$}

%%%%%%%%%%%%%%%%%%%%%%%%%%%%%%%%%%%%%%%%%%%%%%%%%%%%%%%%%%%%%%%%%%%%%%%
% BEGIN FIGURE CODE %%%%%%%%%%%%%%%%%%%%%%%%%%%%%%%%%%%%%%%%%%%%%%%%%%%
%%%%%%%%%%%%%%%%%%%%%%%%%%%%%%%%%%%%%%%%%%%%%%%%%%%%%%%%%%%%%%%%%%%%%%%
%
{\fontsize{6}{6}\selectfont\begin{tikzpicture}[scale = \myfactor]

  \pgfdeclarelayer{bg}     % declare background layer
  \pgfsetlayers{bg,main}   % set the order of the layers (main is the standard layer)

  % Draw input to GNN
  \node (input) [rectangle, minimum width = 0.1*\unit] {$\mathbf{x}$};
  %%%%%%%%%%%%%%%%%%%%%%%%%%%%%%%%%%%%%%%%%%%%%%%%%%
  % Layer 1 %%%%%%%%%%%%%%%%%%%%%%%%%%%%%%%%%%%%%%%%
  %%%%%%%%%%%%%%%%%%%%%%%%%%%%%%%%%%%%%%%%%%%%%%%%%%
  % Draw Layer 1 Filter block and Feature block
  \path (input.east)      ++ \deltainput node [filter]       (L1 Filter1) {\one};
  \path (L1 Filter1) ++ \deltasigma node [nonlinearity] (L1 F1)      {\sigmaone};
  \path (L1 F1) ++ \deltasigma node [pooling] (L1 F2)      {\proyone};
  % Draw connectors for Layer 1
  %\path[draw, -stealth] (L1 Filter1.east) -- node [above] {$\mathbf{z}_1$} (L1 F1.west);
  \path[draw, -stealth] (L1 Filter1.east) -- node [above] {$\mathbf{y}_1$} (L1 F1.west);
  \path[draw, -stealth] (L1 F1.east) -- node [above] {$\mathbf{z}_1$} (L1 F2.west);

  %%%%%%%%%%%%%%%%%%%%%%%%%%%%%%%%%%%%%%%%%%%%%%%%%%
  % Layer 2 %%%%%%%%%%%%%%%%%%%%%%%%%%%%%%%%%%%%%%%%
  %%%%%%%%%%%%%%%%%%%%%%%%%%%%%%%%%%%%%%%%%%%%%%%%%%
  % Draw Layer 2 Filter block and Feature block
  \path (L1 Filter1) ++ (0,-\deltalayer) node [filter]       (L2 Filter1) {\two};
  \path (L2 Filter1) ++ \deltasigma      node [nonlinearity] (L2 F1)      {\sigmatwo};
  \path (L2 F1) ++ \deltasigma node [pooling] (L2 F2)      {\proytwo};
  % Draw connectors for Layer 2
  \path[draw, -stealth] (L2 Filter1.east) --  node [above] {$\mathbf{y}_2$} (L2 F1.west);
  \path[draw, -stealth] (L2 F1.east) -- node [above] {$\mathbf{z}_2$} (L2 F2.west);  
  
  %%%%%%%%%%%%%%%%%%%%%%%%%%%%%%%%%%%%%%%%%%%%%%%%%%
  % Layer 3 %%%%%%%%%%%%%%%%%%%%%%%%%%%%%%%%%%%%%%%%
  %%%%%%%%%%%%%%%%%%%%%%%%%%%%%%%%%%%%%%%%%%%%%%%%%%
  % Draw Layer 2 Filter block and Feature block
  \path (L2 Filter1) ++ (0,-\deltalayer) node [filter]       (L3 Filter1) {\three};
  \path (L3 Filter1) ++ \deltasigma      node [nonlinearity] (L3 F1)      {\sigmathree};
  \path (L3 F1) ++ \deltasigma node [pooling] (L3 F2)      {\proythree};
  % Draw connectors for Layer 2
  %\path[draw, -stealth] (L3 Filter1.east) --  node [above] {$\mathbf{z}_3$} (L3 F1.west);
  \path[draw, -stealth] (L3 Filter1.east) --  node [above] {$\mathbf{y}_3$} (L3 F1.west);
  \path[draw, -stealth] (L3 F1.east) -- node [above] {$\mathbf{z}_3$} (L3 F2.west); 

  %%%%%%%%%%%%%%%%%%%%%%%%%%%%%%%%%%%%%%%%%%%%%%%%%%
  % Connections across layers %%%%%%%%%%%%%%%%%%%%%%
  %%%%%%%%%%%%%%%%%%%%%%%%%%%%%%%%%%%%%%%%%%%%%%%%%%
  % From input to layer 1
  \path[draw, -stealth] (input.east) -- (L1 Filter1.north);
  % From layer 1 to layer 2
  \path (L1 F2.south) ++ (0,-\deltaconnector) node [] (aux1) {};
  \path[draw, -stealth] (L1 F2.south) -- node [below right] {$\mathbf{x}_1$} (aux1.north) 
                                      --                         (aux1.north -| L2 Filter1.north) 
                                      -- node [above left]  {$\mathbf{x}_1$} (L2 Filter1.north);
  % From layer 2 to layer 3
  \path (L2 F2.south) ++ (0,-\deltaconnector) node [] (aux1) {};
  \path[draw, -stealth] (L2 F2.south) -- node [below right] {$\mathbf{x}_2$} (aux1.north) 
                                      --                         (aux1.north -| L2 Filter1.north) 
                                      -- node [above left]  {$\mathbf{x}_2$} (L3 Filter1.north);
  % From layer 3 to output
  \path[draw, -stealth] (L3 F2.south) -- ++ \deltaoutput -- ++ (0.5, 0) 
                        node [right]{$\mathbf{x}_3$};

  %%%%%%%%%%%%%%%%%%%%%%%%%%%%%%%%%%%%%%%%%%%%%%%%%%
  % Light shades to signify layer groups   %%%%%%%%%
  %%%%%%%%%%%%%%%%%%%%%%%%%%%%%%%%%%%%%%%%%%%%%%%%%%
  \begin{pgfonlayer}{bg} 
      % Shading for Layer 1  
      \path (L1 Filter1.west |- L1 F1.south) ++ (-0.4,-0.7)
           node [filter, anchor = south west,
                 fill = black!5, 
                 %minimum width  = 10.75*\unit,
                 minimum width  = 11.8*\unit,
                 minimum height = 2.1*\unit,] 
        (layer)
        {}; 
       \path (layer.south west) ++ (0.0,0.0) node [above right] {$(\ccalA_1, \ccalM_1, \rho_1)$};
      % Shading for Layer 2  
      \path (L1 Filter1.west |- L2 F1.south) ++ (-0.4,-0.7)
           node [filter, anchor = south west,
                 fill = black!5, 
                 %minimum width  = 10.75*\unit,
                 minimum width  = 11.8*\unit,
                 minimum height = 2.1*\unit,] 
        (layer)
        {}; 
       \path (layer.south west) ++ (0.0,0.0) node [above right] {$(\ccalA_2, \ccalM_2, \rho_2)$};
      % Shading for Layer 3  
      \path (L1 Filter1.west |- L3 F1.south)  ++ (-0.4,-0.7)
           node [filter, anchor = south west,
                 fill = black!5, 
                 %minimum width  = 10.75*\unit,
                 minimum width  = 11.8*\unit,
                 minimum height = 2.1*\unit,] 
        (layer)
        {}; 
       \path (layer.south west) ++ (0.0,0.0) node [above right] {$(\ccalA_3, \ccalM_3, \rho_3)$};  \end{pgfonlayer}

\end{tikzpicture}} 
	\caption{(Adapted from~\cite{algnn_nc_j}) Example of an Algebraic Neural Network  with three layers. In the $\ell$-layer, information is first processed by a convolutional filter, then a pointwise nonlinearity, $\eta_{\ell}$, and finally by a pooling operator, $P_{\ell}$.
 The input signal $\mathbf{x}$ is mapped by the network into $\mathbf{x}_{3}$. }
	\label{fig_6}
\end{figure}
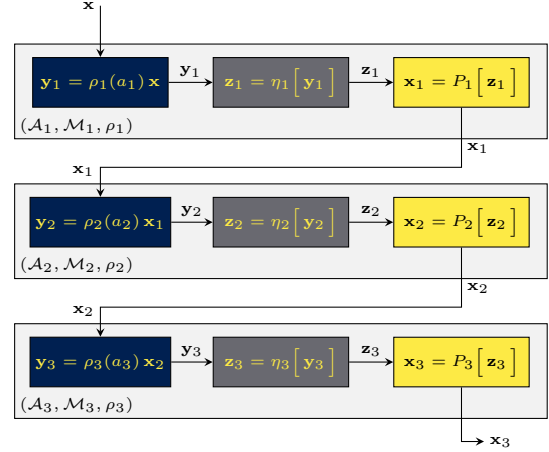

%%---- End of Figure  -----------

Algebraic neural networks (AlgNNs) are layered structures that generalize convolutional neural networks from an algebraic perspective~\cite{algnn_nc_j,parada_algnn}. In each layer, information is processed by means of an algebraic convolutional filter, a point-wise nonlinearity and a pooling operator -- see Fig.~\ref{fig_6}. Then, if the ASM in the $\ell$-th layer of an AlgNN is $(\ccalA_{\ell}, \ccalM_{\ell}, \rho_{\ell})$, the output in that layer can be written as
\begin{equation}\label{eq:xl}
	\bbx_{\ell}=\sigma_{\ell}\left(\rho_{\ell}(a_{\ell})\bbx_{\ell-1}\right)=\Phi (\mathbf{x}_{\ell-1},\mathcal{P}_{\ell-1},\mathcal{S}_{\ell-1}),
\end{equation}
where $\sigma_{\ell}$ is the composition between the pointwise nonlinearity, $\eta_{\ell}$, and the pooling operator, $P_{\ell}$. $\ccalP_{\ell}$ indicates a subset of filters, and $\ccalS_{\ell}$ is the set of admissible shift operators. We use the symbol $\Phi (\mathbf{x}_{\ell-1},\mathcal{P}_{\ell-1},\mathcal{S}_{\ell-1})$ to emphasize that the filters used in the processing of the signal $\bbx_{\ell-1}$ belong to the subset $\ccalP_{\ell}\subset\ccalA_{\ell}$. The map associated with the whole AlgNN is  $\Phi\left(\mathbf{x},\{ \mathcal{P}_{\ell} \}_{1}^{L},\{ \mathcal{S}_{\ell}\}_{1}^{L}\right)$ acting on an input signal $\bbx$. We will use $\left\lbrace\left(\mathcal{A}_{\ell},\mathcal{M}_{\ell},\rho_{\ell};\sigma_\ell \right)\right\rbrace_{\ell=1}^{L}$ to denote an AlgNN with $L$ layers, emphasizing the convolutional model used in each layer, $(\ccalA_{\ell}, \ccalM_{\ell}, \rho_{\ell})$, and the map $\sigma_{\ell}$.

%%%%%%%%%%%%%%%%%%%%%%%%%%%%%%%%%%%%%%%%%%%%%%%
%%%%%%%%%%%%% PERTURBATIONS AND STABILITY %%%%%%%%%%%%%%%
%%%%%%%%%%%%%%%%%%%%%%%%%%%%%%%%%%%%%%%%%%%%%%%

%!TEX root = ../conf_paper.tex

%%%%%%%%%%%%%%%%%%%%%%%%%%%%%%%%%%%%%%%%%%%%%%%%%%%%%%%%%%%%%%%%%%%%%
%%%%%%%%%%%%%%%%%%   S   E   C   T   I   O   N   %%%%%%%%%%%%%%%%%%%%
%%%%%%%%%%%%%%%%%%%%%%%%%%%%%%%%%%%%%%%%%%%%%%%%%%%%%%%%%%%%%%%%%%%%%

\section{Perturbations in Non Commutative Convolutional 
 Models}\label{sec:perturbandstability}

The homomorphism, $\rho$, in any ASM, $(\ccalA,\ccalM,\rho)$, instantiates or implements the filters in the algebra. Therefore, it is natural to consider perturbations as deformations on the homomorphism~\cite{algnn_nc_j,parada_algnn}. In this context we say that $(\ccalA,\ccalM,\widetilde{\rho})$ is a perturbed version of $(\ccalA,\ccalM, \rho)$, with the action of $\widetilde{\rho}$ given by
\begin{equation}\label{eqn_def_perturbation_model_10}
\tdrho(p(g_1, \ldots, g_{m})) 
       = p\big(
               \tilde{g}_1, 
                   \ldots, 
               \tilde{g}_m    
           \big) 
        = p\big(
               \tilde{\bbS}_1, 
                   \ldots,
               \tilde{\bbS}_{m} 
            \big)
            ,
\end{equation}
where $\tilde{\bbS}_i = \bbS_i + \bbT (\bbS_i)$ and $\bbT(\bbS_i)$ is a small diffeomorphism in the space of operators, such that the perturbation on the homomorphism acts as a perturbation on the shift operators. For our discussion, we consider perturbations of the form
\begin{equation}\label{eqn_perturbation_model_absolute_plus_relative}
   \bbT(\bbS_i)=\bbT_0+ \bbT_{1}\bbS_i,
\end{equation}
where the $\bbT_i$ are fixed compact normal operators with $\Vert\bbT_{i}\Vert \ll1$ and 
$
\left\Vert 
\bbT_{i}
\right\Vert_{F}
\leq
\delta
\left\Vert 
\bbT_{i}
\right\Vert 
,   
$
with $\delta>0$.

When considering the perturbation of an AlgNN, we will use  
$\left\lbrace\left( \mathcal{A}_{\ell},\mathcal{M}_{\ell},\tilde{\rho}_{\ell} ;\sigma_\ell \right)\right\rbrace_{\ell=1}^{L}$ to denote the perturbed version of $\left\lbrace\left(\mathcal{A}_{\ell},\mathcal{M}_{\ell},\rho_{\ell};\sigma_\ell \right)\right\rbrace_{\ell=1}^{L}$. In this way, we emphasize that we consider perturbations for all convolutional operators in the network.

%%%%%%%%%%%%%%%%%%%%%%%%%%%%%%%%%%%%%%%%%%%%%%%%%%%
%%%%%%%%%%%%%%%%   S   E   C   T   I   O   N   %%%%%%%%%%%%%%%%%%%%%%%
%%%%%%%%%%%%%%%%%%%%%%%%%%%%%%%%%%%%%%%%%%%%%%%%%%%

\section{Stability Results}\label{sec:stabilitytheorems}

Taking into account the notion of perturbation/deformation introduced before, we now introduce the definition of stability considered under our analysis.

%%------------------------------------------------------------------
%%------------- D   E   F   I   N   I   T   I   O   N --------------
%%------------------------------------------------------------------

\begin{definition}[\cite{algnn_nc_j,parada_algnn}]\label{def:stabilityoperators1} Given filters $p(\mathbf{S}_{1}, \ldots, \bbS_{m})$ and $p(\tilde{\mathbf{S}}_{1}, \ldots, \tilde{\bbS}_{m})$ defined on the models $(\ccalA,\ccalM,\rho)$ and $(\ccalA,\ccalM,\tdrho)$ respectively, we say that $p$ is stable if there exist constants $C_{0}, C_{1}>0$ such that 
\begin{multline}\label{eq:stabilityoperators1}
\left\Vert
      p(\mathbf{S}_1, \ldots, \bbS_m )\mathbf{x} 
      - 
      p(\tilde{\mathbf{S}}_1, \ldots, \tilde{\bbS}_m)\mathbf{x}
\right\Vert
\leq
\\
\left[
C_{0} \sup_{\bbS_{i}\in\ccalS}\Vert\mathbf{T}(\mathbf{S}_i)\Vert + C_{1}\sup_{\bbS_{i}\in\ccalS}\big\|D_{\bbT}(\bbS_i)\big\|
+\mathcal{O}\left(\Vert\mathbf{T}(\mathbf{S}_i)\Vert^{2}\right)
\right] \big\| \bbx \big\|,
\end{multline}
for all $\bbx\in\ccalM$. In \eqref{eq:stabilityoperators1}, $D_{\bbT}(\bbS)$ is the Fr\'echet derivative of the perturbation operator $\bbT$. 
\end{definition}

%%------------------------ End of Definition  ----------------------

To formally state the stability results, we need to first introduce  definitions regarding subsets of filters in the algebra. To do so, we exploit the Fourier representation of the filters. We say that a filter $p(g_1, \ldots, g_m)\in\ccalA$ with Fourier representation $p(x_1, \ldots, x_m)$ is $L_0$-Lipschitz if
\begin{multline}
\left\Vert
p(x_{1},\ldots,x_{m})-p(\tilde{x}_{1},\ldots,\tilde{x}_{m})
\right\Vert  
\\
\leq
L_{0}\left\Vert 
(x_{1},\ldots,x_{m})-(\tilde{x}_{1},\ldots,\tilde{x}_{m})
\right\Vert,                
\end{multline}
for all $x_{i}, \tilde{x}_{i}$. Additionally, we say that $p(g_1,\ldots,g_m)$ is $L_1$-integral Lipschitz if there exists $L_{1}>0$ such that
\begin{equation}
\left\Vert 
\bbD_{p\vert x_{i}}(x_{1},\ldots,x_{m})
\left\lbrace \left(\cdot\right)x_{i}\right\rbrace   
\right\Vert
\leq L_{1}
,
\end{equation}
$\forall~x_{i}$, where $\bbD_{p\vert x_{i}}(x_{1},\ldots,x_{m})$ is the partial Fr\'echet derivative of $p(x_{1},\ldots,x_{m})$, and $\Vert \cdot\Vert$ is the operator norm. We will denote by $\ccalA_{L_{0}}$ and $\ccalA_{L_{1}}$ the sets of $L_0 -$Lipschitz and $L_1$-integral Lipschitz filters in $\ccalA$, respectively.

With these results at hand, we are ready to introduce our first stability result.

%%-------------------------------------------------
%%---------------- COROLLARY ----------------------
%%-------------------------------------------------

\begin{theorem}[\cite{algnn_nc_j}]\label{corollary:stabmultgenfilt}
	Let $(\ccalA,\ccalM,\rho)$ be a non commutative ASM where $\ccalA$ has generators $\{ g_{i}\}_{i=1}^{m}$. Let $(\ccalA,\ccalM,\tilde{\rho})$ be a perturbed version of $(\ccalA,\ccalM,\rho)$ associated to the perturbation model in~(\ref{eqn_perturbation_model_absolute_plus_relative}). If $p\in\mathcal{A}_{L_{0}}\cap\mathcal{A}_{L_{1}}\subset\ccalA$, the operator $p(\bbS_1,\ldots,\bbS_m)$ is stable in the sense of Definition~\ref{def:stabilityoperators1} with $C_{0}=m\delta L_{0}$ and $C_{1}=m\delta L_{1}$.
\end{theorem}
\begin{proof} 
See~\cite{algnn_nc_j}.
\end{proof}

%%------------- End of COROLLARY ------------------

Theorem~\ref{corollary:stabmultgenfilt} shows that one can find stable filters in non commutative convolutional models, and that stability is linked to the selectivity of the filters. More specifically, a reduction in the magnitude of $L_0$ and $L_1$ implies a reduction in the values of the stability constants, i.e. the filter is more stable. However, low values of $L_0$ and $L_1$ imply that the variability of the filter is reduced. In particular, a small value of $L_1$ forces the filter to be flat for large frequencies. From these observations, we can see that although the non commutativity of the signal model substantially changes the nature of the Fourier representation of the filters (matrix functions instead of scalar functions), there is still a trade-off between selectivity and stability analog to that observed for commutative models.

Now, we state a stability result for the operator for an arbitrary layer of any non commutative algebraic neural network.

%%----------------------------------------------------------
%%-------------- THEOREM STABILITY ALGNN 1 -------------------
%%----------------------------------------------------------

\begin{theorem}[\cite{algnn_nc_j}]\label{theorem:stabilityAlgNN0}
	
Let $\left\lbrace\left(\mathcal{A}_{\ell},\mathcal{M}_{\ell},\rho_{\ell};\sigma_\ell \right)\right\rbrace_{\ell=1}^{L}$ be an algebraic neural network and $\left\lbrace\left( \mathcal{A}_{\ell},\mathcal{M}_{\ell},\tilde{\rho}_{\ell} ;\sigma_\ell \right)\right\rbrace_{\ell=1}^{L}$ its perturbed version by means of the perturbation model in~(\ref{eqn_perturbation_model_absolute_plus_relative}). We consider one feature per layer and non commutative algebras $\mathcal{A}_{\ell}$ with $m$ generators. If  $\Phi\left(\mathbf{x}_{\ell-1}, \mathcal{P}_{\ell},\mathcal{S}_{\ell}\right)$ and 
$\Phi\left(\mathbf{x}_{\ell-1},\mathcal{P}_{\ell},\tilde{\mathcal{S}}_{\ell}\right)$ represent the $\ell$-th mapping operators of the AlgNN and its perturbed version, it follows that
\begin{multline}
\left\Vert
               \Phi\left(\mathbf{x}_{\ell-1},\mathcal{P}_{\ell},\mathcal{S}_{\ell}\right)
                -
               \Phi\left(\mathbf{x}_{\ell-1},\mathcal{P}_{\ell},\tilde{\mathcal{S}}_{\ell}\right)
\right\Vert
                 \leq
 \\
                        C_{\ell}\delta
                             \Vert
                                    \mathbf{x}_{\ell-1}
                             \Vert
                             m
                             \left(
                                     L_{0}^{(\ell)} \sup_{\bbS_{i,\ell}}\Vert\bbT^{(\ell)}(\bbS_{i,\ell})\Vert 
                             \right.  
                             \\
                             \left.      
                                     +
                                     L_{1}^{(\ell)}\sup_{\bbS_{i,\ell}}\Vert D_{\mathbf{T^{(\ell)}}}(\bbS_{i,\ell})\Vert
                            \right)
                            ,
\label{eq:theoremstabilityAlgNN0}
\end{multline}
where $C_{\ell}$ is the Lipschitz constant of $\sigma_{\ell}$, and $\mathcal{P}_{\ell}=\mathcal{A}_{L_{0}}\cap\mathcal{A}_{L_{1}}$ represents the domain of $\rho_{\ell}$. The index $\ell$ makes reference to quantities and constants associated to the layer $\ell$.

\end{theorem}
\begin{proof}
See~\cite{algnn_nc_j}.
\end{proof}

%%-------------------- End of THEOREM ----------------------

From Theorem~\ref{theorem:stabilityAlgNN0}, we can observe that the map $\sigma_{\ell}$, i.e. the composition between the point-wise nonlinearity and the pooling operator, scales the stability bound derived for the filters but does not affect its functional structure. In fact, if one chooses $C_{\ell}=1$, the bound is essentially the same. However, as shown in~\cite{algnn_nc_j}, $\sigma_{\ell}$ maps information between different frequency bands. That is, $\sigma_\ell$ does not affect the stability attributes of the filters, but does enrich how information is processed between layers. 

Now, we present the stability result for the operators that entirely characterizes any non commutative AlgNN.

%%--------------------------------------------------------
%%--------------- THEOREM STABILITY ALGNN ------------------
%%--------------------------------------------------------

\begin{theorem}\label{theorem:stabilityAlgNN1}

Let $\left\lbrace\left(\mathcal{A}_{\ell},\mathcal{M}_{\ell},\rho_{\ell};\sigma_\ell \right)\right\rbrace_{\ell=1}^{L}$ be an algebraic neural network and $\left\lbrace\left( \mathcal{A}_{\ell},\mathcal{M}_{\ell},\tilde{\rho}_{\ell} ;\sigma_\ell \right)\right\rbrace_{\ell=1}^{L}$ its perturbed version by means of the perturbation model in~(\ref{eqn_perturbation_model_absolute_plus_relative}). We consider one feature per layer and non commutative algebras $\mathcal{A}_{\ell}$ with $m$ generators. If  $\Phi\left(\mathbf{x},\{ \mathcal{P}_{\ell} \}_{1}^{L},\{ \mathcal{S}_{\ell}\}_{1}^{L}\right)$ and 
$\Phi\left(\mathbf{x},\{ \mathcal{P}_{\ell} \}_{1}^{L},\{ \tilde{\mathcal{S}}_{\ell}\}_{1}^{L}\right)$ represent the mapping operator 
and its perturbed version, it follows that 
\begin{multline}
\left\Vert
\Phi
      \left(
              \mathbf{x},\{ \mathcal{P}_{\ell} \}_{1}^{L},\{ \mathcal{S}_{\ell}\}_{1}^{L}
      \right)
-
\Phi
       \left(
              \mathbf{x},\{ \mathcal{P}_{\ell} \}_{1}^{L},\{ \tilde{\mathcal{S}}_{\ell}\}_{1}^{L}
       \right)
\right\Vert
\\
\leq
\sum_{\ell=1}^{L}\boldsymbol{\Delta}_{\ell}\left(\prod_{r=\ell}^{L}C_{r}\right)\left(\prod_{r=\ell+1}^{L}B_{r}\right)
\left(\prod_{r=1}^{\ell-1}C_{r}B_{r}\right)\left\Vert\mathbf{x}\right\Vert
,
\label{eq:theoremstabilityAlgNN1}
\end{multline}
where $C_{\ell}$ is the Lipschitz constant of $\sigma_{\ell}$, $\Vert\rho_{\ell}(a)\Vert\leq B_{\ell}~\forall~a\in\mathcal{P}_{\ell}$, and $\mathcal{P}_{\ell}=\mathcal{A}_{L_{0}}\cap\mathcal{A}_{L_{1}}$ represents the domain of $\rho_{\ell}$. The functions $\boldsymbol{\Delta}_{\ell}$ are given by
\begin{equation}\label{eq:varepsilonl}
\boldsymbol{\Delta}_{\ell}
           =\delta m\left(
           L_{0}^{(\ell)} \sup_{\bbS_{i,\ell}}\Vert\bbT^{(\ell)}(\bbS_{i,\ell})\Vert 
           \right.
           %\\
           \left.
          +
          L_{1}^{(\ell)}\sup_{\bbS_{i,\ell}}\Vert D_{\mathbf{T^{(\ell)}}}(\bbS_{i,\ell})\Vert
             \right)
             ,
\end{equation}
with the index $\ell$ indicating quantities and constants associated to the layer $\ell$.
\end{theorem}
\begin{proof} See~\cite{algnn_nc_j}. \end{proof}

%%--------------------- End of THEOREM -------------------

One of the fundamental observations from Theorem~\ref{theorem:stabilityAlgNN1} is that the AlgNN can be stable to perturbations, and that stability is inherited from the convolutional filters. This is expected since the maps $\sigma_{\ell}$ only scale the stability bounds derived for the filters. However, it is important to remark that while the stability bounds derived for filters, layer operators and AlgNNs are essentially the same (with appropriate normalization of $C_\ell$ and $B_\ell$~\cite{algnn_nc_j}), the discriminability power of each of these operators is substantially different. This is one of the aspects that explains why convolutional networks perform better than filter banks~\cite{algnn_nc_j,parada_algnn}. 
Due to the fact that the AlgNN is enriched with point-wise nonlinearities for the same level of stability, AlgNNs are more stable than the corresponding convolutional filters. Additionally, notice that the redistribution of information in the frequency domain performed by $\sigma_{\ell}$ occurs by mapping information between spaces of dimension larger than one.

%%%%%%%%%%%%%%%%%%%%%%%%%%%%%%%%%%%%%%%%%%%%%%%
%%%%%%%%%%%%%  %%%%%%%%%%%%%%%
%%%%%%%%%%%%%%%%%%%%%%%%%%%%%%%%%%%%%%%%%%%%%%%

%\input{sec_SpectOP.tex}
%
%
%%-------------------------------------------------------------
%%-----------SECTION: PROOFS OF THEOREMS
%%--------------------------------------------------------------
%\input{sec_proofTheorems.tex}
%
%
%%----------------------------------------------------------
%%---------SECTION: DISCUSSION -----------------------------
%%----------------------------------------------------------
%\input{sec_discussion.tex}
%
%
%
%%%%%%%%%%%%%%%%%%%%%%%%%%%%%%%%%%%%%%%%%%%%%%%
%%%%%%%%%%%%% Numerical Experiments %%%%%%%%%%%%%%%
%%%%%%%%%%%%%%%%%%%%%%%%%%%%%%%%%%%%%%%%%%%%%%%

%!TEX root = ../conf_paper.tex

%
%%%%%%%%%%%%%%%%%%%%%%%%%%%%%%%%%%%%%%%%%%%%%%%%%%%%%%%%%%%%%%%%%%%%%%%%%%%%%%%%
%%%   S   E   C   T   I   O   N   %%%%%%%%%%%%%%%%%%%%%%%%%%%%%%%%%%%%%%%%%%%%%%
%%%%%%%%%%%%%%%%%%%%%%%%%%%%%%%%%%%%%%%%%%%%%%%%%%%%%%%%%%%%%%%%%%%%%%%%%%%%%%%%
%
\section{Numerical Experiments}\label{sec_Numerical_Experiments}

In this section, we demonstrate the stability properties of the non commutative filters employed in multigraph neural networks \cite{msp_j, msp_icassp2023}.  Under the multigraph signal model, which is associated with the regular algebra of polynomials with multiple independent variables, the existence of subclasses of filters that are Lipschitz and Integral Lipschitz are guaranteed (Theorem~\ref{corollary:stabmultgenfilt}). Therefore, it is possible to learn filters that can mitigate the effect of deformations on the shift operators. To do so, we train an architecture to learn filters using a cost function that adds a penalty based on the filter's Integral Lipschitz constants, and showcase that this model is much more resilient to perturbations enforced on the graph. 

We consider the application of multigraph learning for a rating prediction task on the Jester dataset \cite{goldberg2001eigentaste}. For our multigraph model, we consider two classes of edges connecting the jokes: one representing sentence similarity and the other modeling rating similarity amongst users in the training set. The multigraph is sparsified by keeping only the 20 strongest edges of each class per node. We arbitrarily selected the first joke for evaluating the performance of our predictions.

For our experiment, we compare the performance of a learned linear multigraph filter (MFilter), a multigraph neural network (MGNN), and another which is regularized by an estimate of the filter’s Integral Lipschitz constant (MGNN IL). All architectures are composed of a convolutional layer with 3 filter taps and a readout layer mapping to a rating estimate. We report the root mean squared error (RMSE) and the standard deviation after training for 40 epochs across 5 random splits of the dataset.

\begin{figure}[t]
\includegraphics[width=\linewidth]{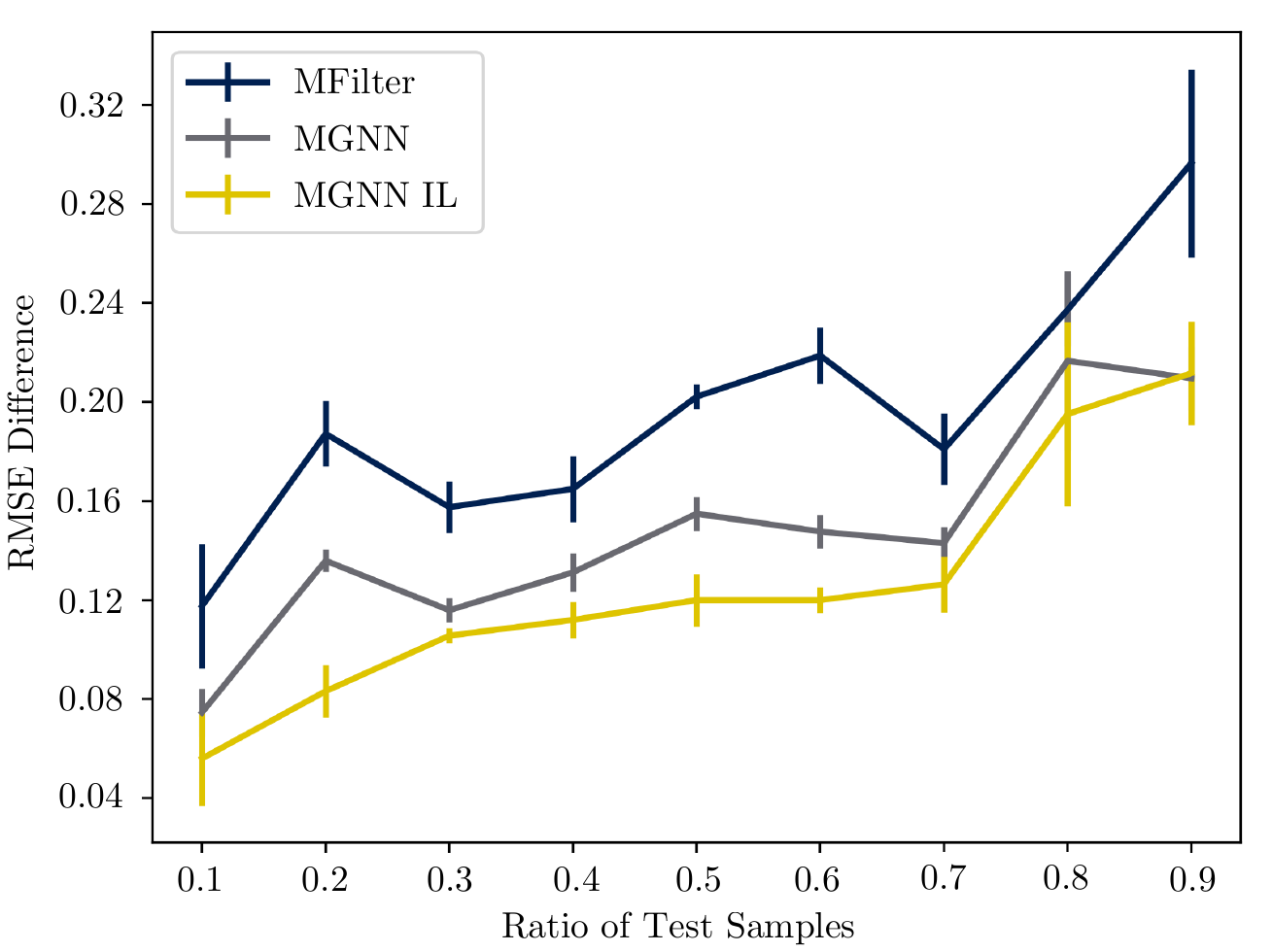}
\centering
\caption{Results of the stability experiment.}
\label{fig:multiData-one}
\end{figure}

To simulate stability towards estimation error of the shift operators, we first train each architecture on 90\% of the training set. At evaluation, we replace the rating similarity shift operator with an estimate generated by a random subset of the training set ranging from 10\% to 90\% of the overall dataset and compare the difference in RMSE between the trained and evaluated models. As can be seen in Figure \ref{fig:multiData-one}, the penalized MultiGNN consistently maintains the smallest difference in error for each size of the test set, demonstrating the most stability towards estimation error. 

%%----------------------------------------------------------
%%---------SECTION: CONCLUSIONS ----------------------------
%%----------------------------------------------------------
%!TEX root = conf_paper.tex

%%
%%----------------------------------------------------------------------------
%%-------- SECTION: CONCLUSIONS --------------------------------- ------------
%%----------------------------------------------------------------------------
%%
%%------------------- Title ------------------------------
\section{Conclusions}\label{sec:conclusions}
\vspace{-2mm}

We leveraged algebraic signal processing (ASP) to model general non commutative convolutional filters and showed that neural networks built upon these models can be stable to deformations. The results obtained apply to existing architectures such as multigraph-CNNs, quaternion-CNNs, quiver-CNNs and non commutative group-CNNs. We proved that non commutative convolutional neural networks can be stable, which is an attribute inherited from the stability of the filters. This is due to the fact that the inclusion of pointwise nonlinearities and pooling operators only change the stability bounds derived for the filters by a scalar factor. The differences between the Fourier representations of non commutative and commutative filters lead to stability conditions that are different for both scenarios. While in commutative signal models the stability conditions are imposed on scalar valued functions, in non commutative models such conditions are imposed on matrix polynomials. We demonstrated that the polynomial representations in non commutative models exhibit a similar trade-off between stability and selectivity.

%
%
%

%
%\section{REFERENCES}
%\label{sec:refs}
%

% References should be produced using the bibtex program from suitable
% BiBTeX files (here: strings, refs, manuals). The IEEEbib.bst bibliography
% style file from IEEE produces unsorted bibliography list.
% -------------------------------------------------------------------------
\clearpage
\bibliographystyle{unsrt}
\bibliography{./bibliography}
\clearpage

\end{document}